\documentclass{article}
\usepackage{arxiv}
\usepackage{xspace}
\usepackage{subfigure}
\usepackage{graphicx}
\usepackage{booktabs}
\usepackage{amsmath}
\usepackage{amssymb}
\usepackage{color}
\usepackage{url}

\def\BibTeX{{\rm B\kern-.05em{\sc i\kern-.025em b}\kern-.08emT\kern-.1667em\lower.7ex\hbox{E}\kern-.125emX}}
    
\newcommand{\eat}[1]{}

\newcommand{\ie}{\textit{i.e.,}\xspace}
\newcommand{\eg}{\textit{e.g.,}\xspace}
\newcommand{\etal}{\textit{et al.}\xspace}
\newcommand{\emb}{Obj-GloVe\xspace}
\newcommand{\paratitle}[1]{\noindent\textbf{#1}}

\newcommand{\red}[1]{\textcolor{red}{#1}}

\begin{document}

%
\title{\emb: Scene-Based Contextual Object Embedding}

\author{
Canwen Xu \\
  Wuhan University\\
  Hubei, China \\
  \texttt{xucanwen@whu.edu.cn} \\
   \And
Zhenzhong Chen\thanks{Corresponding author.} \\
  Wuhan University\\
  Hubei, China \\
  \texttt{zzchen@whu.edu.cn} \\
  \And
  Chenliang Li \\
  Wuhan University\\
  Hubei, China \\
  \texttt{cllee@whu.edu.cn} \\
}

\maketitle

\begin{abstract}
Recently, with the prevalence of large-scale image dataset, the co-occurrence information among classes becomes rich, calling for a new way to exploit it to facilitate inference. In this paper, we propose Obj-GloVe, a generic scene-based contextual embedding for common visual objects, where we adopt the word embedding method GloVe to exploit the co-occurrence between entities. We train the embedding on pre-processed Open Images V4 dataset and provide extensive visualization and analysis by dimensionality reduction and projecting the vectors along a specific semantic axis, and showcasing the nearest neighbors of the most common objects. Furthermore, we reveal the potential applications of Obj-GloVe on object detection and text-to-image synthesis, then verify its effectiveness on these two applications respectively. 
\end{abstract}

\section{Introduction}
Recently, the size of image datasets is continuously increasing, bringing unprecedented research opportunity and inspiring new ways to exploit rich implicit information in images. With the development of Deep Learning, visual features are more and more well-used by Deep Convolutional Neural Network (DCNN) based methods. However, we wonder if it is possible to draw an inference by scene understanding. Getting inspiration from Natural Language Processing (NLP), we find that contextual embedding, which aims to represent entities in low-dimensional vector space, is perfect for capturing implicit relations among visual objects. More importantly, similar to word embedding in NLP, a contextual embedding for visual objects can enable many downstream applications. To formally define, a contextual object embedding is an embedding learned to represent visual objects based on their contexts.

Embedding has many remarkable characteristics. First of all, the distance between two entities can be easily measured by calculating the cosine or Euclidean distance between their vector representations. Moreover, a vector representation can be decomposed or projected, which is useful to research latent relations between entities. Although word embedding has its own limitation that possible meanings of words are conflated into a single representation, our object embedding is not affected by this defect since visual objects are unambiguous. This is because the authors of image datasets have disambiguated the possible meanings for one particular class when designing the label classes (\eg in Open Images, ``keyboard'' is disambiguated to be ``computer keyboard'' and ``musical keyboard'').

When designing the embedding model for visual objects, GloVe \cite{emnlp14:pennington} comes into our minds at once. GloVe exploits the co-occurrence between entities, which has already been proved workable in Computer Vision (CV) by early studies \cite{nips09:malisiewicz,cacm10:torralba,iccv07:rabinovich}. Also, GloVe is fast to train and effective in many NLP tasks. Thus, we propose \textbf{\emb} (\textbf{Obj}ect \textbf{Glo}bal \textbf{Ve}ctors), a generic scene-based object embedding. We process the images first to extract the labels to convert it to a language-like corpus, which contains image-converted ``sentences''. Then we train the embedding on this corpus to get the final representation for each object. In this process, no information from pixels is used. Note that we deliberately separate visual features from our embedding since there have already been many effective methods to encode visual features (\eg ResNet \cite{cvpr16:he}, VGG \cite{iclr15:simonvan}). Thus, a future computer vision system may benefit from combining embedding-based relation inference with traditional visual features. Also, pre-existing models can benefit from \emb with minimal modification.

We carefully make the choice of which dataset to train the embedding on. In our work, we choose Open Images V4 \cite{openimages} mainly considering its scale. Open Images V4 has 600 hierarchical classes, while COCO \cite{mscoco} and PASCAL-VOC \cite{pascalvoc}, the previously most used object detection datasets, have only 80 and 20 object classes, respectively. Open Images covers all classes in COCO and PASCAL-VOC which means our embedding can be directly used on COCO and PASCAL-VOC by constructing a mapping dictionary. Also, previous research \cite{iis16:lai} shows that the corpus size is critical for embedding training. Open Images V4 has 1,910,098 images (compared to 200,000 in COCO and 11,530 in PASCAL-VOC), making it the largest object detection dataset ever.

\emb is rather conceptual and pioneering, but it is also practical and capable of augmenting current CV systems. The application of \emb is two-fold. First, the rich semantics contained in \emb can serve as side information for computer vision systems to help to deal with situations where visual features cannot be used. For example, when the object detection system suffer from low confidence when classifying an object, \emb can help the recognition by analyzing its context. Furthermore, when an object is invisible due to terrible weather or being blocked by other objects, \emb can help to infer the object from the scene. We call such application \textbf{object inference}. Besides, the entries of \emb are language-based (\ie object labels), containing similar semantics (\eg co-occurrence, word analogy) like word embedding, which makes it wonderful for \textbf{bridging the language and vision}. Since objects are directly fixed to labels, \emb implicitly trains natural language and images in a joint vector space in essence. This advantage can be exploited in language-related CV tasks. Moreover, it provides possible way to better exploit language priori knowledge to solve CV problems.

In this paper, we propose \emb, a generic contextual object embedding based on co-occurrence.
To the best of our knowledge, we are the first to propose an embedding for common visual objects.
We provide detailed visualization and analysis on \emb. We would like to highlight its impressive performance on gender axis projection.
Also, we show potential applications in object detection and image-to-text synthesis enrichment by involving object inference and bridging language and vision.

\section{{\emb}}
To exploit the scene-based co-occurrence within an image, we realize co-occurrence probability is critical since it effectively measures how strong the relationship is between two classes. We adopt GloVe (\textbf{Glo}bal \textbf{Ve}ctors for Word Representation) embedding~\cite{emnlp14:pennington}, which exploits co-occurrence probability and is famous for its excellent performance in NLP. We convert each annotated image in training set to a ``sentence'' then apply \emb on all converted ``sentences'' to get a scene-based embedding. In this section, we first introduce our pre-processing step, then give an introduction to \emb and how we train it.

\subsection{Data Pre-processing}
\begin{figure}[h]
  \centering
  \includegraphics[width=12cm]{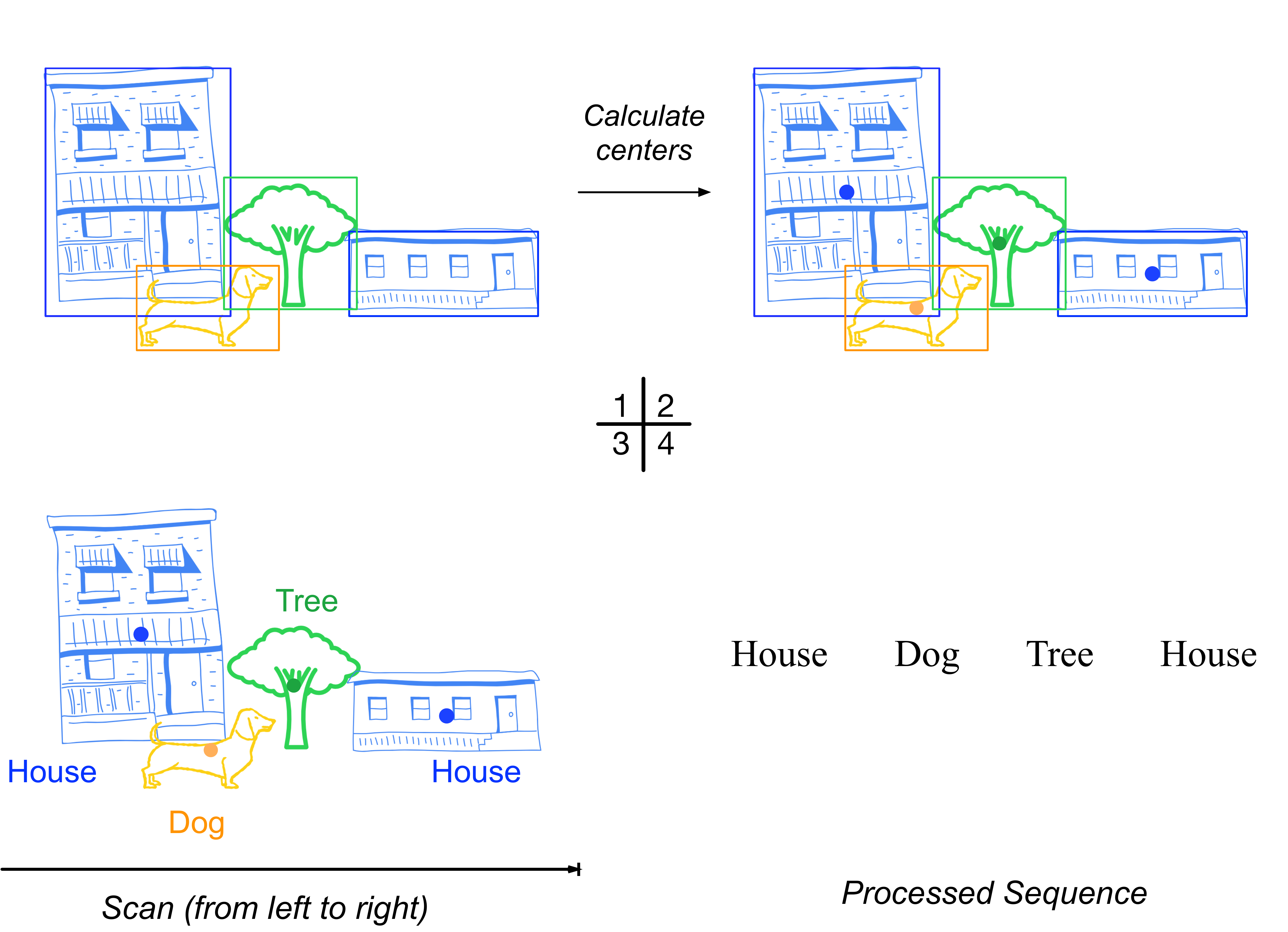}
  \caption{The procedure of data pre-processing. (1)(2) We calculate the center of bounding boxes. (3)(4) Then we scan the image horizontally to get the object label sequence.}
  \label{fig:preprocess}
\end{figure}
The pre-processing is illustrated in Figure \ref{fig:preprocess}. First, we calculate the center of each bounding box by:
\begin{equation}
    C = (\frac{x_1 + x_2}{2}, \frac{y_1 + y_3}{2})
\end{equation}
where $(x_1, y_1)$, $(x_2, y_2)$, $(x_3, y_3)$, and $(x_4, y_4)$ are the coordinates of the four anchors of the bounding box, clockwise.
Then, we scan the bounding boxes horizontally or vertically to get the sequence of objects. Due to the aspect ratio of most images in the dataset, we apply a horizontal scan in our pre-processing step.

\subsection{Glove Embedding}
GloVe was proposed by Pennington \etal \cite{emnlp14:pennington} to better exploit the statistical co-occurrence information in the text. GloVe has soon been widely accepted as an effective word representation and used in many downstream tasks of NLP. We adopt GloVe by putting it into the context of visual object embedding, on the basis of our pre-processed data.

Formally, let the matrix of object-object co-occurrence counts be denoted by $X$, whose entry $X_{ij}$ tabulates the number of times that object $j$ occurs in the same image (\ie context) of object $i$. Let $X_i = \sum_k X_{ik}$ be the number of the times any object appears in the same image of object $i$. Finally, let $P_{ij} = P(j|i) = X_{ij}/X_i$ be the probability that word $j$ appear in the same image of object $i$.

Then, the ratios of co-occurrence probabilities are taken into consideration of learning process. The ration $P_{ik}/P_{jk}$ depends on three words, $i$, $j$, and $k$. Generally, the model (denoted as $F$) takes the form,
\begin{equation}
    F \left( o _ { i } , o _ { j } , \tilde { o } _ { k } \right) = \frac { P _ { i k } } { P _ { j k } }
    \label{equ:f1}
\end{equation}
where $o_* \in \mathbb { R } ^ { d }$ are object vectors and $\tilde{o}_* \in \mathbb{R}^d$ are separate context object vectors. In this equation, $P_{ik}/P_{jk}$ is extracted from the corpus (\ie the set of processed annotations). Note that only co-occurrence within a specific window size $w$ is counted. Since the vector space is linear by nature, $F$ can depend only on the difference of two target objects, modifying Equation \ref{equ:f1} to be:
\begin{equation}
    F \left( o _ { i } - o _ { j } , \tilde { o } _ { k } \right) = \frac { P _ { i k } } { P _ { j k } }
\end{equation}
Since the right side is a scalar, $F(\cdot)$ should also output a scalar, which makes the equation to be:
\begin{equation}
    F \left( \left( o _ { i } - o _ { j } \right) ^ { T } \tilde { o } _ { k } \right) = \frac { P _ { i k } } { P _ { j k } }
\end{equation}
Since an object and a context object are arbitrary, which makes both $o\leftrightarrow\tilde{o}$ and $X\leftrightarrow X^T$ interchangeable. To represent $P$ with $F$ function, the equation is modified to be:
\begin{equation}
    F \left( \left( o _ { i } - o _ { j } \right) ^ { T } \tilde { o } _ { k } \right) = \frac { F \left( o _ { i } ^ { T } \tilde { o } _ { k } \right) } { F \left( o _ { j } ^ { T } \tilde { o } _ { k } \right) }
\end{equation}
in which,
\begin{equation}
\label{equ:f5}
    F \left( o _ { i } ^ { T } \tilde { o } _ { k } \right) = P_{ik} = \frac{X_{ik}}{X_i}
\end{equation}
The solution to Equation \ref{equ:f5} is $F=exp$, or, 
\begin{equation}
    o _ { i } ^ { T } \tilde { o } _ { k } = \log \left( P _ { i k } \right) = \log \left( X _ { i k } \right) - \log \left( X _ { i } \right)
\end{equation}
In this equation, it would exhibit the exchange symmetry if $\log(X_i)$ is not on the right side. Note that this term is independent of $k$ which means it can be absorbed into a bias $b_i$ for object $i$. To restore the symmetry, an additional bias $\tilde{b}_k$ is added for $\tilde{o}_k$,
\begin{equation}
\label{equ:f7}
    o _ { i } ^ { T } \tilde { o } _ { k } + b _ { i } + \tilde { b } _ { k } = \log \left( X _ { i k } \right)
\end{equation}

Equation \ref{equ:f7} is a simplification of Equation \ref{equ:f1} but ill-defined since the logarithm diverges whenever its argument is zero. To resolve that, $\log(X_{ik})$ is replaced by $\log(1+X_{ik})$, which avoids the divergences while maintaining the sparsity of $X$.
A downside is that the model weighs all co-occurrences equally including rare or non-existent pairs. However, such rare co-occurrences are noisy and less meaningful than frequent ones.

To address the problem, a new weighted least squares regression model is proposed. After introducing a weighting function $f(X_{ij})$ into to the cost function, the model has the form:
\begin{equation}
\label{equ:j}
	J=\sum_{i, j=1}^{V} f\left(X_{i j}\right)\left(o_{i}^{T} \tilde{o}_{j}+b_{i}+\tilde{b}_{j}-\log X_{i j}\right)^{2}
\end{equation}
where $V$ is the number of the object classes. The weighting function obey following constraints:
 \begin{enumerate}
  \item $f(0)=0$. If $f$ is regarded as a continuous function, it should vanish as $x\rightarrow 0$ fast enough that the $\lim _{x \rightarrow 0} f(x) \log ^{2} x$ is finite.
  \item $f(x)$ should be non-decreasing to ensure rare co-occurrences not overweighted.
  \item $f(x)$ should be relatively small for large values of $x$ to ensure frequent co-occurrences not overweighted either.
\end{enumerate}

A class of functions working well was proposed by Pennington \etal, which can be parameterized as:
\begin{equation}
\label{equ:f}
	f(x)=\left\{\begin{array}{cl}{\left(\frac{x}{x_{\max }}\right)^{\alpha}} & {,\text { if } x<x_{\max }} \\ {1} & {,\text { otherwise }}\end{array}\right.
\end{equation}

We use the same parameters as the ones taken in the original paper, which are $100$ for $x_{max}$ and $3/4$ for $\alpha$.

\subsection{Training}
We train our embedding on 14,610,229 box labels in 1,743,042	 images on the training set of Open Images V4 \cite{openimages}. The Open Images V4 includes 600 classes and we discard object classes which appear less than $10$ times in the training set to be $596$ entries at last. We select the window size $w$ according to an experiment in Section \ref{sec:od}.

\begin{table*}
\centering
  \caption{The nearest neighbors of the most common entries of \emb. The number following each neighbor is cosine distance between the entry and the neighbor.}
  \label{tab:neighbors}
  \begin{tabular}{lrrrr}
    \toprule
    Object&Count&\#1 &\#2&\#3 \\ 
    \midrule
    Clothing&1,438,128&Man (0.484)& Human Face (0.516)& Woman (0.548) \\
    Man&1,418,594&Clothing (0.484)& Human Face (0.542)& Footwear (0.576) \\
    Tree&1,051,344&Plant (0.553)& Skyscraper (0.624)& Wheel (0.627)\\ 
    Human Face&1,037,710&Glasses (0.376)& Woman (0.440)& Girl (0.442) \\
    Person & 1,034,721&Clothing (0.550)& Man (0.679)& Footwear (0.683)\\
    Woman & 767,337 & Girl (0.317)& Human face (0.440)& Dress (0.511)\\
    Footwear & 744,474 & Jeans (0.529)& Shorts (0.572)& Human leg (0.576)\\
    Window & 503,467 & House (0.316)& Building (0.412)& Door (0.418)\\
    Flower & 345,296 & Flowerpot (0.385)& Rose (0.401)& Houseplant (0.430)\\
    Wheel & 340,639 & Tire (0.147)& Car (0.313)& Bicycle Wheel (0.346)\\
    Plant & 267,913 & Tree (0.533)& Flower (0.614) & Duck (0.644) \\ 
    Car & 248,075 & Land Vehicle (0.303) & Wheel (0.313) & Truck (0.380) \\

  \bottomrule
\end{tabular}
\end{table*}

\begin{figure}[h]
  \centering
  \includegraphics[width=12cm]{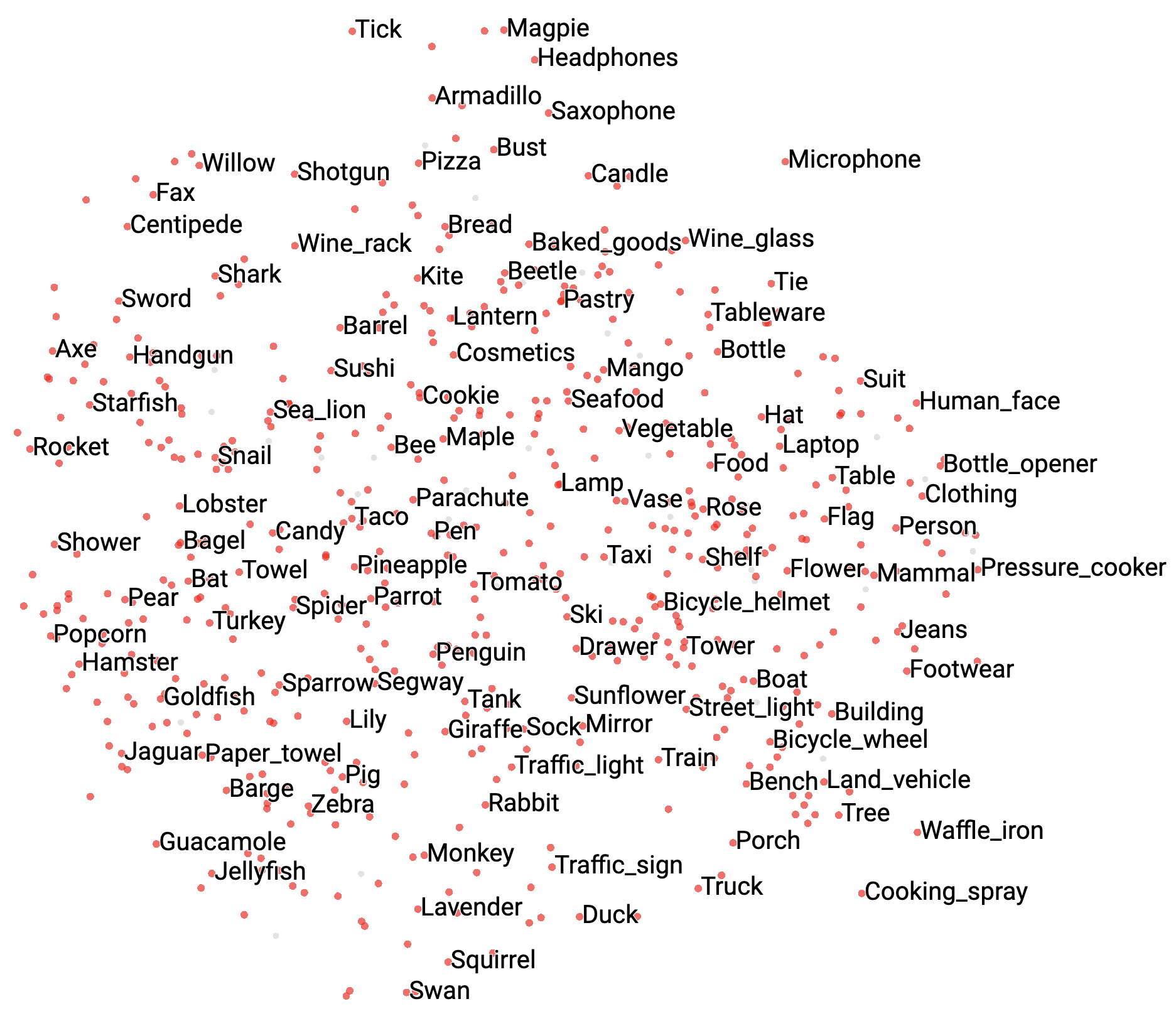}
  \caption{PCA visualization of \emb. The labels are randomly sampled.}
  \label{fig:pca}
\end{figure}

\begin{figure}[h]
  \centering
  \includegraphics[width=12cm]{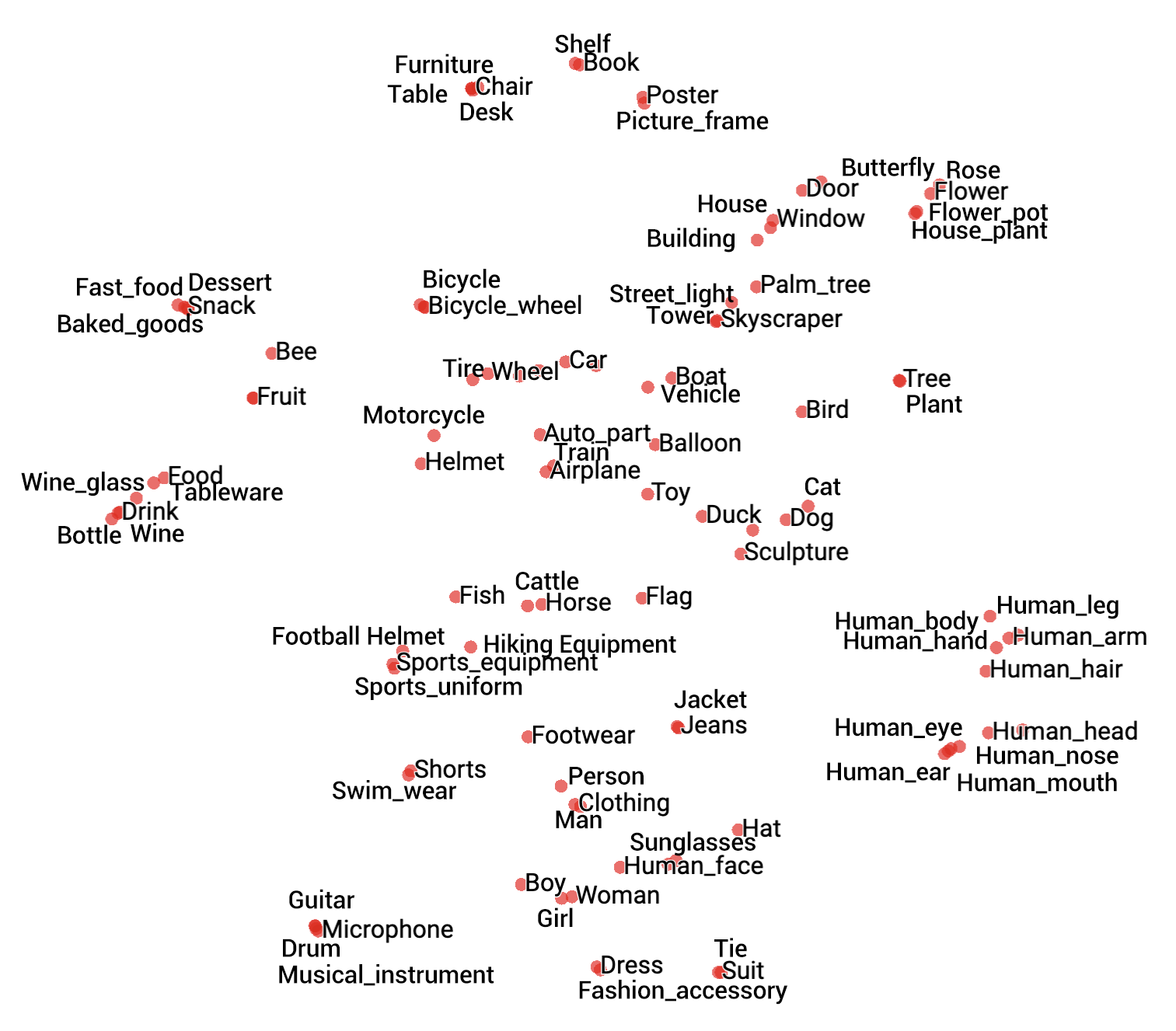}
  \caption{t-SNE visualization of the most common objects of \emb.}
  \label{fig:tsne}
\end{figure}

\begin{figure}[h]
  \centering
  \includegraphics[width=12cm]{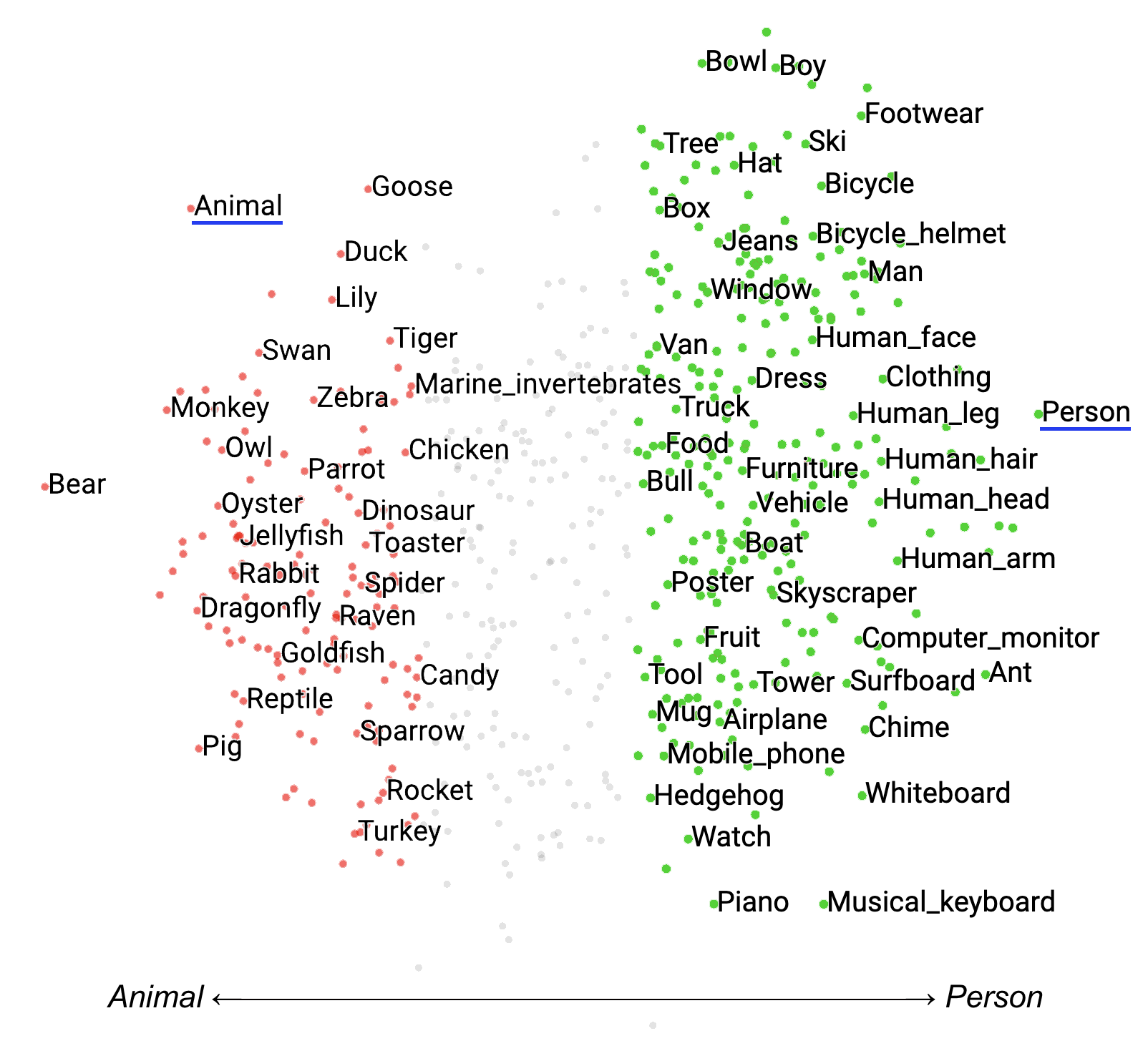}
  \caption{Projections of \emb on \textit{Animal-Person} axis. The projections closer to \textit{Animal} and \textit{Person} are marked with red and green, respectively. The labels are randomly sampled.}
  \label{fig:project}
\end{figure}

\begin{figure}[h]
  \centering
  \includegraphics[width=\linewidth,height=1.17\linewidth]{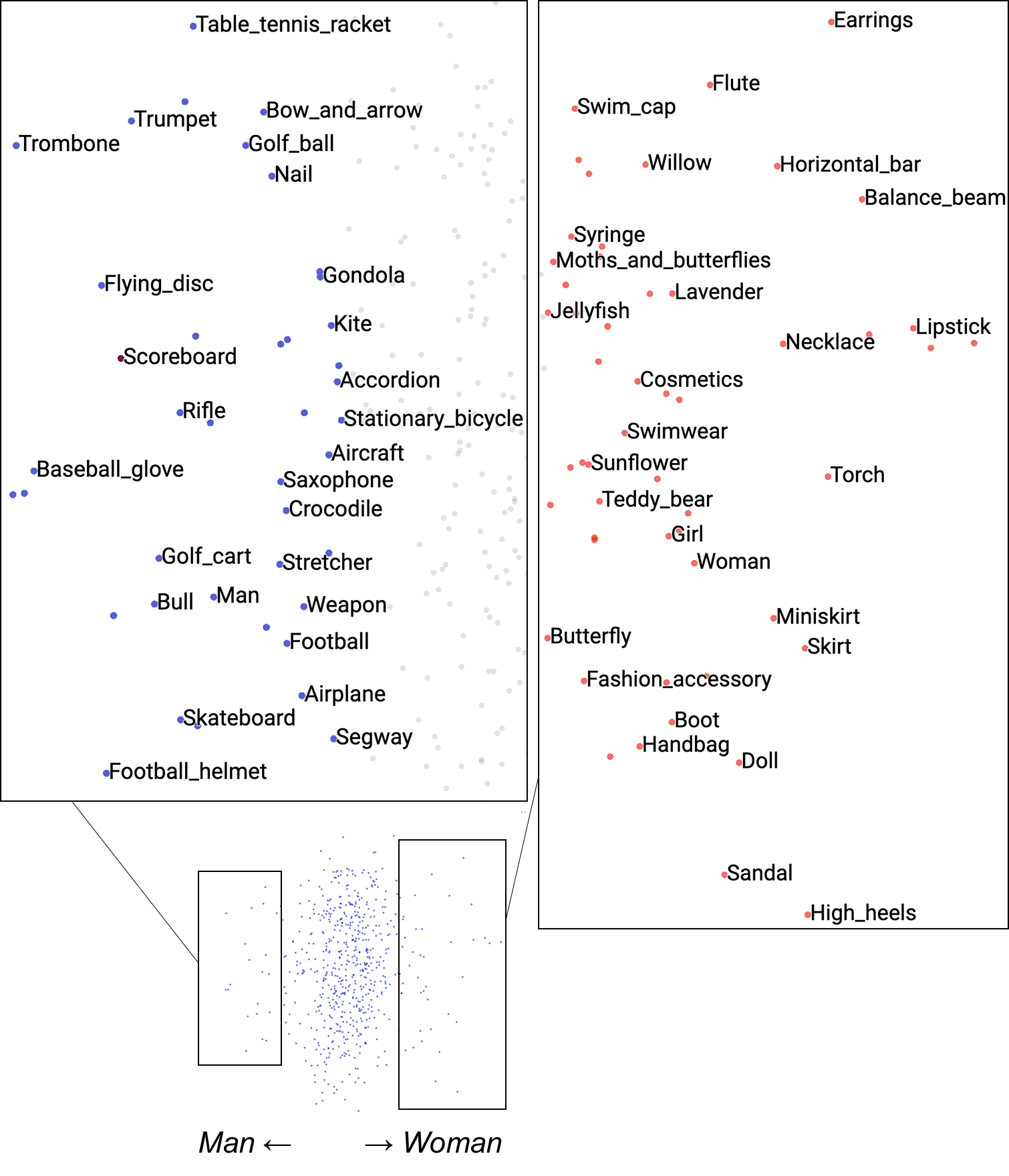}
  \caption{Projections of \emb on \textit{Man-Woman} axis. We visualize the pole areas of two sides. The labels are randomly sampled.}
  \label{fig:gender}
\end{figure}

\begin{figure}[h]
  \centering
  \includegraphics[width=12cm]{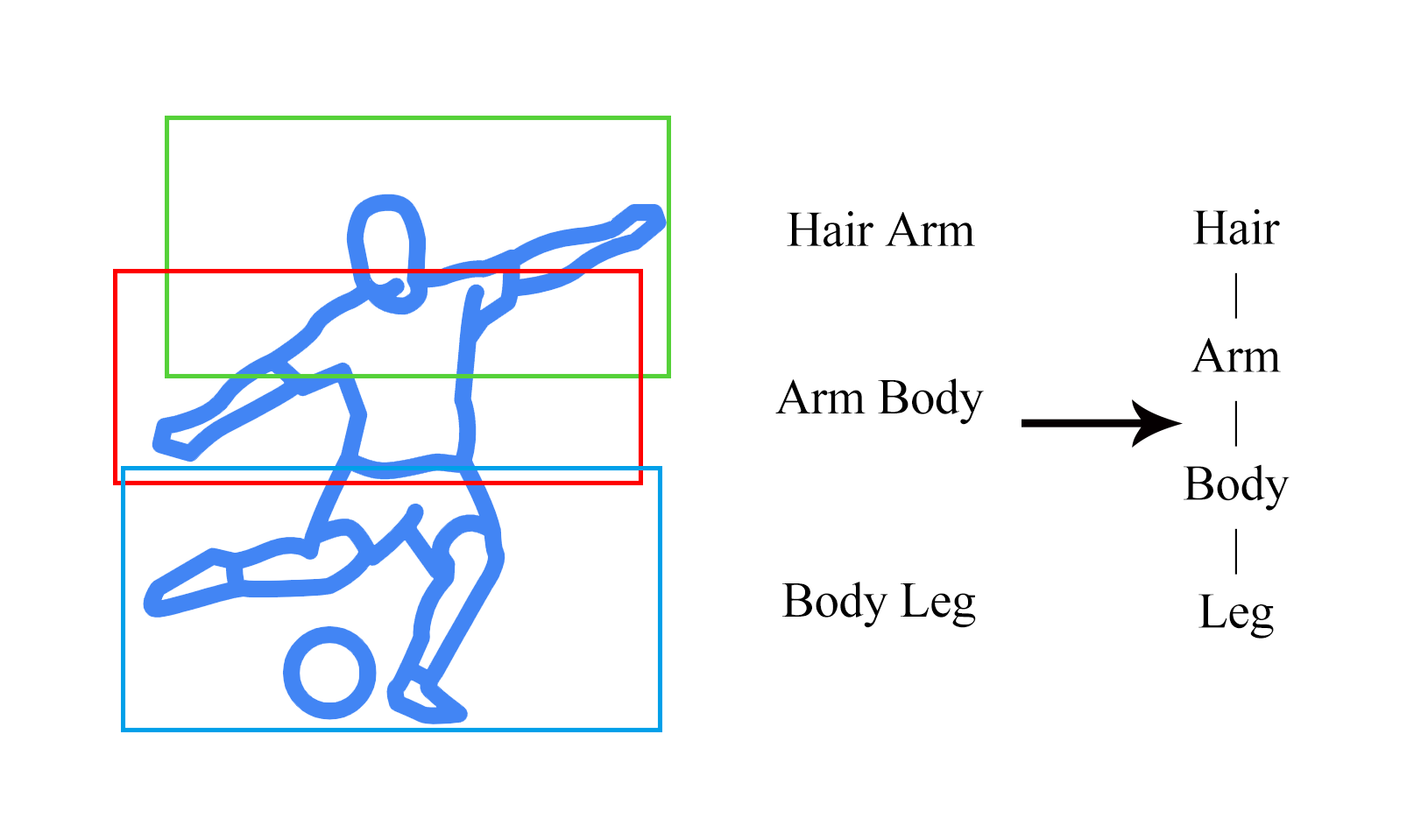}
  \caption{Photos are slices of a full scene. Thus, \emb can implicitly exploit spatial information across multiple images of the same scene during training.}
  \label{fig:slice}
\end{figure}

\section{Visualization and Analysis}
\label{sec:vis}
\paratitle{Dimensionality Reduction.}
To visualize \emb, we use Principal Component Analysis (PCA) to reduce the dimensionality to 2D.
The visualization result is shown in Figure \ref{fig:pca}. Also, shown in Figure \ref{fig:tsne}, we visualize the most common objects (appearing more than 10,000 times in the training set) with T-distributed Stochastic Neighbor Embedding (t-SNE), giving an impressive result.
In this figure, we can see several main clusters based on scenes.
To be detailed, the scene of music performance is at the bottom left of the figure, including entries \textit{Guitar}, \textit{Microphone}, \textit{Drum}, and \textit{Musical Instrument}.
The indoor scene is at the top of the figure, including various furniture. The dining scene is at the left of the figure, including tableware and food.
The portrait scene is at the bottom right including human parts. 
The clothing, the house and decoration, and sports scene are at the bottom, top right, and middle left, respectively. The outdoor scene, which mainly includes vehicles, is at the middle of the figure.

It may seem, at first glance, that \emb does not contain any positional information. However, showing at the bottom right of Figure \ref{fig:tsne}, the human parts are displayed in the spatial order (\ie leg-body-arm-hair). In fact, shown in Figure \ref{fig:slice}, \emb is capable of implicitly capturing positional information. Since a photo can be seen as a spatial slice of a full scene, the co-occurrence probability is larger between objects which are spatially closer. Ideally, spatial relation can be learned well across multiple images if the training set is large enough.

\paratitle{Projection.}
Due to the nature of vector space, a good embedding should contain precise semantics which performs well on vector decomposition and projection.
Shown in Figure \ref{fig:project}, we project all vectors on the axis of \textit{Animal-Person}, making \textit{Animal} on the left while \textit{Person} on the right.

Recently, many studies have focused on gender bias in social media. Since all images in Open Images V4 dataset are from Flickr, a photo sharing social media, it is possible to research implicit gender bias and stereotyping with \emb. Thus, we project the objects along \textit{Man-Woman} axis. The result is shown in Figure \ref{fig:gender}. Although most objects are ``neutral'' (\ie locating in the middle of the axis), there are still some objects highly relevant to gender. 

\paratitle{Nearest Neighbors.}
We list the most common entries and their nearest five neighbors in the vector space in Table \ref{tab:neighbors}. Note that in this table, we filter out entries appearing less than $10,000$ times in the training set to concentrate on the most common objects and exclude entries which may be underfitted caused by lack of samples (we will discuss this in Section \ref{limitations}).

In general, to be close in the vector space, two entries should meet one of the following conditions: (1) they refer to the same or similar objects (\eg \textit{Person}, \textit{Woman}, and \textit{Man}). (2) they are a common co-occurring pair (\eg \textit{Glasses} and \textit{Human Face}).

\section{Applications}
In this section, we provide two application instances - \textbf{object detection} and \textbf{text-to-image synthesis enrichment} for \textbf{object inference} and \textbf{language bridging}, respectively.

\begin{figure}[h]
  \centering
  \includegraphics[width=12cm,height=6.336cm]{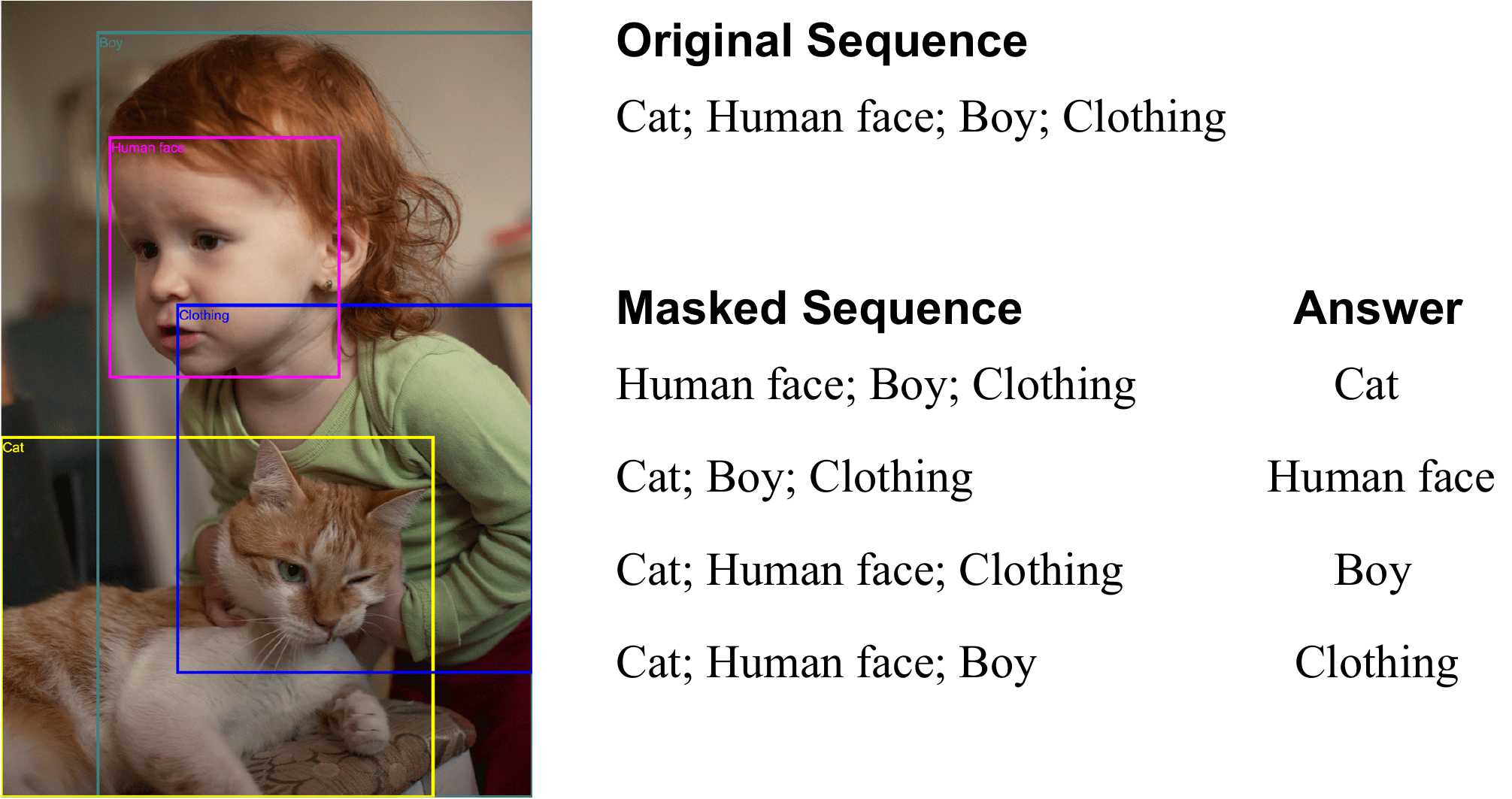} 
  \caption{A live example of masking. The labels in the image are iteratively masked to form a test sample. Thus, the original sequence is converted into masked sequences and answers, which are the input and output of the task, respectively.}
  \label{fig:mask}
\end{figure}

\begin{table}
\centering
  \caption{The results of \emb with different window sizes versus random guessing on Open Images V4 validation set.}
  \label{tab:masking}
  \begin{tabular}{lrrr}
    \toprule
    Window Size&Acc@1&Acc@5&Acc@10\\
    \midrule
    Random Guessing & 0.17& 0.83& 1.66\\
    \hline
    5 & 34.70 & 61.99 & 69.40\\
    10 & \textbf{36.33} & 69.06 & \textbf{81.04}\\
    15 & 35.47 & \textbf{69.14} & 80.96\\
  \bottomrule
\end{tabular}
\end{table}

\begin{figure}[h]
  \centering
  \subfigure[A skyscraper.]{
  \includegraphics[width=3.8cm]{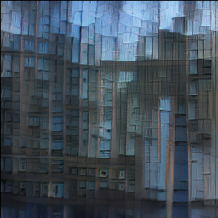}
  \label{subfig:skyscraper}
  }
  \subfigure[A skyscraper \red{with tower}.]{
  \includegraphics[width=3.8cm]{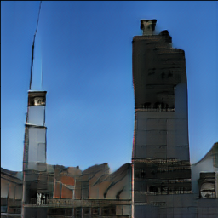}
  \label{subfig:tower}
  }
  \subfigure[A computer monitor.]{
  \includegraphics[width=3.8cm]{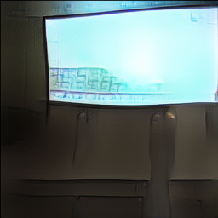}
  \label{subfig:monitor}
  }
  \subfigure[A computer monitor \red{with desk}.]{
  \includegraphics[width=3.8cm]{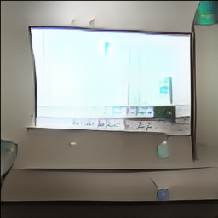}
  \label{subfig:monitordesk}
  }
  \caption{Scene generation with \emb. The black text is the user input and the red one is ``auto-completed'' based on cosine similarity of \emb.}
\end{figure}

\begin{figure*}[t]
  \centering
  \subfigure[A sofa bed.]{
  \includegraphics[width=4cm]{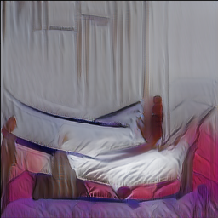}
  }
  \subfigure[A sofa bed \red{with pillow}.]{
  \includegraphics[width=4cm]{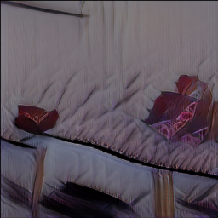}
  }
  \subfigure[A sofa bed \red{with pillow and bookcase}.]{
  \includegraphics[width=4cm]{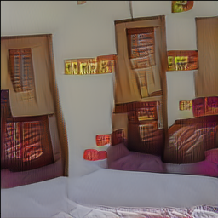}
  }
  \subfigure[A sofa bed \red{with pillow, bookcase and coffee table}.]{
  \label{subfig:sofa4}
  \includegraphics[width=4cm]{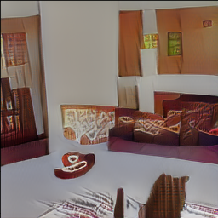}
  }
  \subfigure[A house.]{
  \includegraphics[width=4cm]{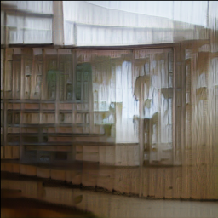}
  }
  \subfigure[A house \red{with window}.]{
  \includegraphics[width=4cm]{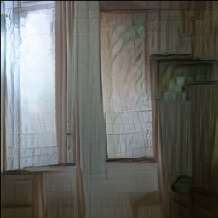}
  }
  \subfigure[A house \red{with window and door}.]{
  \includegraphics[width=4cm]{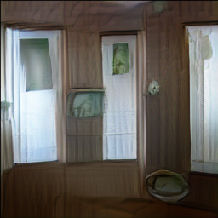}
  }
  \subfigure[A house \red{with window, door, and stairs}.]{
  \label{subfig:house4}
  \includegraphics[width=4cm]{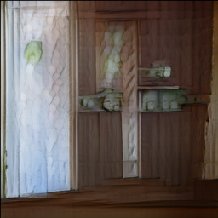}
  }
  \caption{Progressive scene generation with \emb. The black text is the user input and the red one is ``auto-completed'' based on cosine similarity of \emb.}
  \label{fig:scene}
\end{figure*}


\subsection{Object Detection}
\label{sec:od}

Since \emb is trained on object detection dataset, it is natural to apply it ``back''. Before the prevalence of deep learning, context information is widely used to refine object detection \cite{cvpr09:divvala,cvpr11:song,nips09:malisiewicz,cacm10:torralba,iccv07:rabinovich}. Very recently, a few studies \cite{iccv17:chen,cvpr18:hu,cvpr18:liu} have verified the effectiveness of context information in Deep Convolutional Neural Network (DCNN) based object detector.

Containing rich context information, \emb can be exploited by a neural network architecture or a post-processing strategy (\eg Pair Linking~\cite{cikm17:phan,corr18:phan}, which considers both local scores and relations). Furthermore, the small size of \emb ($<600$ KB for 50D vectors) makes it possible to be attached to a mobile object detector and augment its performance by relation inference.

To simply verify the potential of \emb on object detection and set a metric for future research, we define an auxiliary task named ``masking'' to evaluate the performance of contextual object embedding. As illustrated in Figure \ref{fig:mask}, for each annotated image in the validation set with more than one bounding box, we iteratively ``mask'' a box label from the ground truth and attempt to predict it with trained embedding. Note that the visual features are completely discarded in this task. We enumerate all possible classes and sum up the cosine similarity between the embedding of the enumerated class and other labels.
\begin{equation}
    score_i = \sum\limits_{b_j}^{m-1} \text{CosineSimilarity}(b_j, c_i)
\end{equation}
\begin{equation}
	\text { CosineSimilarity}(\mathbf{A},\mathbf{B}) =\frac{\mathbf{A} \cdot \mathbf{B}}{\|\mathbf{A}\|\|\mathbf{B}\|}
\end{equation}
where $m$ is the number of boxes in the image; $score_i$ is the ranking score for i-th class; $b_j$ and $c_i$ are the embedding of the j-th known label and i-th class, respectively.
Then, we predict the blank by simply ranking the scores in descending order.

The results of \emb with different window sizes are shown in Table \ref{tab:masking}. We can see \emb dramatically outperforms random guessing by more than 200 times in terms of accuracy (Acc@1). Although no visual feature is used, the predictive accuracy (Acc@1) achieves $1/3$ and the accuracy at five (Acc@5) exceeds $60\%$. For embedding with a window size of $10$ and $15$, the accuracy at ten (Acc@10) achieves more than $80\%$. To further analyze, a larger window size can capture a wider range of co-occurrence pairs but could be misled by irrelevant pairs (\eg a clock on the desk and the sofa on the other side of the image). In our following experiment, we fix the window size of the embedding to $10$.

\subsection{Text-to-Image Synthesis Enrichment}

Recently, image generation has been well researched. Text-to-image synthesis, which incorporates the text understanding and image generation, has shown its effectiveness~\cite{cvpr18:xu, iclr16:mansimov, cvpr17:nguyen, icml16:reed, iccv17:zhang}. The language-based nature of such models and the ability of multi-object generation are perfect for showing the effectiveness of \emb in bridging language and vision. Also, it verifies that \emb can understand scenes. Here, we experiment with AttnGAN \cite{cvpr18:xu} \footnote{We use the model pre-trained on COCO dataset. \url{https://github.com/taoxugit/AttnGAN}}, the state-of-the-art text-to-image synthesis model. Although current methods can yield discernible images, the generated images often differ greatly from real images in detail. To enrich the details and reconstruct the scene, the object's nearest neighbors are automatically selected from the vector space of \emb and concatenated with the user input object with the preposition ``with''. 

Since \emb is trained over scenes, it has the potential to reconstruct the scene. We start from pairwise enrichment. Shown in Figure \ref{subfig:tower} and \ref{subfig:monitordesk}, the added objects help to put the object back into its context, making the images more realistic.
Shown in Figure \ref{fig:scene}, we start from ``a sofa bed'' and ``a house'' and add the nearest objects in the \emb vector space one by one to progressively generate a full scene. We can see the scene gradually becoming rich in details. In Figure \ref{subfig:sofa4} and \ref{subfig:house4}, although the scene is rather complicated, all four elements in each figure remain recognizable.

%

\section{Related Work}
\paratitle{Word Embedding.} 
Word embedding is a technique in NLP where words are mapped to vectors of real numbers. All word embedding can be classified to language model (LM) based and count-based. LM-based methods attempt to predict the next word with known words. The idea of word embedding becomes popular since embedding is derived as a by-product of Neural Network Language Model (NNLM), the first neural network language model proposed by Bengio \etal \cite{nips00:bengio}. Mikolov \etal \cite{iclr13:mikolov,nips13:mikolov} proposed Skip-Gram and CBOW, which compose Word2Vec, a well-known word embedding method. Different from LM-based methods, count-based embedding methods use statistics to learn representation for each word. Deerwester \etal \cite{jasis90:deerwester} introduced Latent Semantic Analysis (LSA) and Singular Value Decomposition (SVD) applied on a term-document matrix, which can build word embedding. Lund and Burgess \cite{hal} proposed Hyperspace Analogue to Language (HAL), using a context window around the word to get weighted word-word co-occurrence counts to build an co-occurrence matrix. GloVe, proposed by Pennington \etal \cite{emnlp14:pennington}, encodes semantic relationships between words as vector offsets in vector space, exploiting co-occurrence ratios instead of raw co-occurrence count. GloVe is fast to train and outperforms Word2Vec in multiple NLP tasks.

\paratitle{Embedding for Visual Objects.}
The idea of embedding for visual objects is rather new. Schroff \etal \cite{cvpr15:schroff} encoded face features into vector space. Zhang \etal \cite{cvpr17:zhang} proposed a ``Visual Translation Embedding network'' to embed objects in a low-dimensional relation space where a relation is modeled as a simple vector translation (\ie $subject + predicate \approx object$). Wan \etal \cite{ijcai18:wan} proposed an embedding model exploiting both structural and visual features to learn an object and relation embedding on visual relation datasets. The previous work all involve visual features, which is quite different from our work since they are in essence reducing the dimensionality of visual features. To the best of our knowledge, there is currently no work on training a general vector representation for visual objects.

\paratitle{Object Relation in Object Detection.}
Early work used object relations as a post-processing step \cite{cvpr09:divvala,cvpr11:song,nips09:malisiewicz,cacm10:torralba,iccv07:rabinovich}. In these work, the detections are re-scored by considering object relationships. For example, co-occurrence is used by DPM \cite{pami10:felzenszwalb} to refine prediction scores. Also, subsequent studies \cite{pami12:choi,cvpr14:mottaghi} attempt to use more complex relation features, taking position and size into account. Very recently, a few studies verified that modeling relations between objects can also bring improvement to Deep Convolutional Neural Network (DCNN) based object detectors, which is considered to have already implicitly incorporated contextual information. Chen \etal \cite{iccv17:chen} proposed Spatial Memory Network (SMN) for context reasoning in object detection to model instance-level context and successfully improved the performance of Faster-RCNN \cite{nips15:ren} on COCO dataset. Hu \etal \cite{cvpr18:hu} proposed Relation Network for object detection, using a neural network architecture to model relations between objects and forming an end-to-end object detector. Similarly, Liu \etal \cite{cvpr18:liu} proposed a structure inference net using both instance-level relations and scene-level recognition for object detection augmentation.

\paratitle{Text-to-Image Synthesis.} 
The text-to-image synthesis problem is split by Reed \etal \cite{icml16:reed} into two sub-problems: learning a joint embedding between natural language and images and train a deep convolutional generative adversarial network (GAN) to synthesise realistic images. Dong \etal \cite{iccv17:dong} used a pairwise ranking loss to project images and text into a joint embedding space. PPGN \cite{cvpr17:nguyen} exploits a conditional network to restrain the synthetic images on a caption. StackGAN \cite{stackgan} generates realistic images with two stages. DA-GAN \cite{dagan} ``translates'' each word into a region in an image. AttnGAN \cite{cvpr18:xu} harnesses attention mechanism to refine the local details of synthetic images. Current models achieve satisfying performance on datasets of a specific field (\eg CUB dataset \cite{WahCUB_200_2011}) but perform poorly on common objects dataset (\eg COCO).

\section{Limitations and Future Work}
\label{limitations}
\paratitle{Limitations.}
Firstly, we would like to emphasize that our work on object embedding is rather conceptual and experimental. Although we provide two application nuggets, we acknowledge that to make the idea more solid, strong models exploiting object embedding still remain to be proposed and compared to the state-of-the-art solutions in more tasks. Furthermore, the accuracy of \emb is marred by unbalanced object detection dataset with rare classes. In this case, due to the use of weighting function in Equation \ref{equ:j}, these rare classes are ``ignored'' during training which makes their vectors nearly randomized. For example, \textit{Pizza Cutter} appears for only 20 times in the training set and the nearest neighbors of it are \textit{Eraser}, \textit{Pressure Cooker} and \textit{Chisel}, which makes no sense at all. 

\paratitle{Future Work.}
Generally, future work should be two-fold: \textbf{(1) improving the quality of contextual object embedding;
(2) exploring more application of contextual object embedding to improve downstream tasks}.
In our work, we train a generic scene-based representation for objects and attempt to capture the general relations among objects as much as possible. Although as we analyze in Section \ref{sec:vis} that spatial information is implicitly exploited across multiple images, our method is still harsh on pre-processing and ignores the fine-grained spatial relation between two objects. In future work, a graph which contains the spatial information of objects in an image may be constructed and graph-based models (\eg DeepWalk \cite{kdd:perozzi}) may be applied to get the graph embedding.
For the latter, future work may use \emb in object relation task since the co-occurrence information may indicate the presence of object relations (\eg on, in, under, ride). Also, object embedding can be used to bridge the gap between language and vision in more language-related tasks like zero-shot classification, image and video caption, and visual Q{\&}A. Future work may also bring language priori to vision models to improve their performance.

\section{Conclusion}
In this paper, we propose \emb, the first contextual embedding for common visual objects, trained on large-scale object detection dataset. We provide extensive visualization and analysis for the trained embedding. Also, we reveal two applications of \emb on object detection and text-to-image synthesis, showing the effectiveness of \emb on these two applications.

\bibliographystyle{unsrt}
\bibliography{sample-sigconf}

\end{document}